\documentclass[conference]{IEEEtran}
\IEEEoverridecommandlockouts
\usepackage{cite}
\usepackage{amsmath,amssymb,amsfonts}
\usepackage{algorithmic}
\usepackage{graphicx}
\usepackage{textcomp}
\usepackage{xcolor}
\usepackage{fancyhdr}

\def\BibTeX{{\rm B\kern-.05em{\sc i\kern-.025em b}\kern-.08em
    T\kern-.1667em\lower.7ex\hbox{E}\kern-.125emX}}
\makeatletter
\newcommand{\linebreakand}{%
  \end{@IEEEauthorhalign}
  \hfill\mbox{}\par
  \mbox{}\hfill\begin{@IEEEauthorhalign}
}
\makeatother
\makeatletter

\makeatletter
 \let\old@ps@headings\ps@headings
 \let\old@ps@IEEEtitlepagestyle\ps@IEEEtitlepagestyle
 \def\confheader#1{%
 \def\ps@IEEEtitlepagestyle{%
 \old@ps@IEEEtitlepagestyle%
 \def\@oddhead{\strut\hfill#1\hfill\strut}%
 \def\@evenhead{\strut\hfill#1\hfill\strut}%
 }%
 \ps@headings%
 }
 \makeatother

\confheader{%
 2020 23rd International Conference on Computer and Information Technology (ICCIT), 19-21 December, 2020.
 }

 \usepackage[pscoord]{eso-pic}
\newcommand{\placetextbox}[3]{
 \setbox0=\hbox{#3}
 \AddToShipoutPictureFG*{ \put(\LenToUnit{#1\paperwidth},\LenToUnit{#2\paperheight}){\vtop{{\null}\makebox[0pt][c]{#3}}}
 }
 }
 \placetextbox{.23}{0.055}{\small{978-0-7381-2333-2/20/\$31.00 \textcopyright2020 IEEE}}



\let\<\textless
\let\>\textgreater
\usepackage{booktabs}
\usepackage{tabularx}

\begin{document}

\title{Image to Bengali Caption Generation Using Deep CNN and Bidirectional Gated Recurrent Unit\\}

\author{\IEEEauthorblockN{Al Momin Faruk }
\IEEEauthorblockA{\textit{CSE} \\
\textit{American International}\\
\textit{University-Bangladesh}\\
Dhaka,Bangladesh \\
almominfaruk@gmail.com}
\and
\IEEEauthorblockN{Hasan Al Faraby}
\IEEEauthorblockA{\textit{CSE} \\
\textit{American International}\\
\textit{University-Bangladesh}\\
Dhaka,Bangladesh \\
hasanalfaraby.dev@gmail.com}
\and
\IEEEauthorblockN{Md. Muzahidul Azad}
\IEEEauthorblockA{\textit{CSE} \\
\textit{American International}\\
\textit{University-Bangladesh}\\
Dhaka,Bangladesh \\
marufmuzahidul@gmail.com}
\and
\IEEEauthorblockN{Md. Riduyan Fedous}
\IEEEauthorblockA{\textit{CSE} \\
\textit{American International}\\
\textit{University-Bangladesh}\\
Dhaka,Bangladesh \\
mdrfridu786@gmail.com}
\linebreakand
\IEEEauthorblockN{Md. Kishor Morol}
\IEEEauthorblockA{\textit{CSE} \\
\textit{American International University-Bangladesh}\\
Dhaka,Bangladesh \\
kishor@aiub.edu}
}
\maketitle

\begin{abstract}
There is very little notable research on generating descriptions of the Bengali language. About 243 million people speak in Bengali, and it is the 7th most spoken language on the planet. The purpose of this research is to propose a CNN and Bidirectional GRU based architecture model that generates natural language captions in the Bengali language from an image. Bengali people can use this research to break the language barrier and better understand each other's perspectives. It will also help many blind people with their everyday lives. This paper used an encoder-decoder approach to generate captions. We used a pre-trained Deep convolutional neural network (DCNN) called InceptonV3image embedding model as the encoder for analysis, classification, and annotation of the dataset's images Bidirectional Gated Recurrent unit (BGRU) layer as the decoder to generate captions. Argmax and Beam search is used to produce the highest possible quality of the captions. A new dataset called BNATURE is used, which comprises 8000 images with five captions per image. It is used for training and testing the proposed model.  We obtained BLEU-1, BLEU-2, BLEU-3, BLEU-4 and Meteor is 42.6, 27.95, 23, 66, 16.41 28.7, respectively.
\end{abstract}


\section{Introduction}
Caption generating from an image is one of the challenging tasks in the field of computer vision and natural language processing (NLP). It has a considerable impact on different fields of machine learning. For example, it can help visually impaired people better understand the content of images on the web. In many different areas, it can also have favorable impacts, such as human-computer interaction, security, etc.\\
Caption generation models can help solve the computer vision challenge of determining what objects are in an image. However, it has to be powerful enough also to express the relationship between the objects in natural language. Due to these problems, it made caption generation seen as a massive problem for a long time. Nevertheless, there has been significant progress in approaching this problem using end to end approaches\cite{b1}. Many CNN \& RNN based neural network captioning models are trained with parallel image description datasets such as MSCOCO, Flickr30k, Etc. These models have shown impressive performance in the COCO captioning challenge \cite{b2}. According to the automatic metrics, their result may have surpassed human performance in some instances. As it seems, previous research on the caption generating model on the English language has been carried out in great success. It has produced an impressive result in generating English captions. However, there has been very little research conducted which produces captions in the Bengali language. Much research uses machine learning, computer vision, and natural language processing in the Bengali language \cite{b3}\cite{b4}. However, there are no significant works towards building an image captioning system in the Bengali language.\\
This paper follows the latest in machine translation principles known as encoder-decoder \cite{b6}. Which is proven to produce an impressive result in generating proper sentences like \cite{b7}\cite{b8} for image captioning problem. The function of feature extraction of the given image in convolutional neural network (CNN) or ConvNet was used as an encoder. InceptionV3 was used as the architecture of CNN. It has an impressive performance in image annotation \cite{b31}. (GRU) The Gated recurrent unit was used as a decoder, which helped to produce captions from CNN image function extraction input. GRU has been used to overcome some problems which are present in Recurrent Neural Network (RNN). Problems like vanishing gradients have been solved by using GRU instead of RNN \cite{b9}. In our proposed architecture model, we replaced the LSTM layer with GRU for efficiency and better results. As it is similar to a paper proposed by Vinyals et al. \cite{b6}. By following the architecture model, especially in the Bengali Language, "text-to-image generation" works can be easily solved by changing the ConvNet encoder as desired.\\
However, in today's reality performance of end to end captioning models is far from satisfactory according to the metrics based on human judgment \cite{b10}. Therefore, it is sufficient to say that despite the success of the captioning models, this problem is far from being solved.\\
The rest of the paper is organized as follows: In section, II-V we describe our proposed approach for caption generation. Back-End model architecture, Image processing, caption processing and new datasets. Section VI-VIII illustrates caption generations, the evaluation metrics, the results in tabular format and the discussion. Section IX-X concludes the paper with conclusion and references.\\
\subsection{Related Works}
Here we briefly discuss some related works on Convolutional Neural Network (CNN), Gated Neural Network (GRU) and Recurrent Neural network (RNN) and Image captioning, etc.
\subsubsection{Image Captioning}There are lots of image captioning works in English language. But we did not find any reputable research on image captioning in the Bengali language. There are a few recent works on image captioning using the Bengali language like “IMAGETOTEXT” architecture model by  Jishan  et al. cite\cite{b11} where they used Hybrid RNN to produce caption from images. Earlier image captioning approaches are visual primitive recognizer paired with a rule-based sentence generator. There might be another method including a framework using attempted sentences and introducing the scene's triplets into the sentence as in \cite{b12}, Another way is using a phase containing objects and their relationship in an image like in \cite{b13}. All of these rudimentary approaches succeed in generating captions from images nut the output is very rigid and unsatisfactory. Therefore, after Deep convolutional network and end-to-end pattern based caption generation method become available. Many researchers started using CNN based encoder and RNN or LSTM based decoder to generate captions from images \cite{b14}. The generated captions from these approaches are less rigid and more expressive and are many times better than the previous methods.
\subsubsection{Convolutional Neural Network (CNN)} Convolutional Neural network or CNN first became popular in 2012 by Krizhevsk et al. \cite{b15}, he won large scale visual recognition challenge (ILSVRC) in 2012 with ALEXNET architecture model. After that time CNN is used widely in various computer vision problems. CNN is used mainly for classifying images in classes {b16}, detecting objects in images \cite{b17}\cite{b22}, annotating image \cite{b18}etc. There are other reasons that make CNN advance so rapidly. Such as datasets availability (e.g. MSCOCO, BNLIT, Flickr30k) \cite{b6}\cite{b19}\cite{b20}\cite{b11}. Also, the availability of GPU power that helps the model’s process of training faster. CNN is commonly used to address problems with image annotation. CNN solved high-precision image annotation problems. CNN architecture of its latest type, like InceptionV3 was the winner of ILSVRC in 2014 and Runner up by almost 94\% in 2015 for image classification problems with the top 5 accuracy category\cite{b21}.
\subsubsection{Gated Recurrent Unit (GRU)}Gated recurrent Units are gating mechanism of Recurrent Neural Network (RNN). GRU was first introduced in 2014 by Kyunghyun Cho et al.\cite{b7}. It is like the Long short term memory (LSTM) but with a forget gate. It has less parameter than LSTM. It performs similarly with LSTM on certain task like polyphonic music modeling, speech signal modeling and NLP \cite{b23}\cite{b24}. GRU works better than LSTM in smaller and less frequent datasets.\cite{b25}\cite{b26}\cite{b20}. GRU is used in caption decoding instead of  RNN to solve the problem of vanishing gradients. GRU also has a feature that helps it to recall a long series of sequences with other related features.
\section{Approach}
\subsection{Backend Model Architecture}
Deep learning is a part of machine learning which is based on artificial neural network to mimic the cognitive processes of human the brain in the recognition and visual knowledge retrieval. It has become one of the most populous computational techniques in the artificial intelligence nowadays. Among some of the popular deep learning architectures, we have used CNN (Convolution Neural Network) and RNN (Recurrent Neural Network) for producing our captions.
Convolution Neural Network (CNN) consists of convolution layers, which are the building blocks of this artificial neural network. CNN or ConvNet is so far is being extensively used for analyzing images and videos. Although image analysis has been the most popular use, CNN can be used to analyze other data or classification problems as well. We can say CNN has more power to recognize the patterns and sense them. In CNN, there are input layer, convolution layer, pooling layers, fully connected layer, and output layer \cite{b38} shown in Fig. 1. The input layer holds the picture value of the picture as there are three colors channel value holds by this layer. The CONV layer measures the performance of neurons attached to the local input areas, each computing a dot-product between their weights and a tiny area with which input volumes are associated. The input volume is related to the increasing calculation of a dot between its weights and a small component. Learnable filters are also known as the kernel, are the small components of the CONV layer. In ConvNet, pooling layer comes after the CONV layer. The pooling layer gradually decreases the spatial representation size to diminish the number of parameters and network calculation and thereby, governance over fitting. 
\begin{figure}[!b]
\centerline{\includegraphics{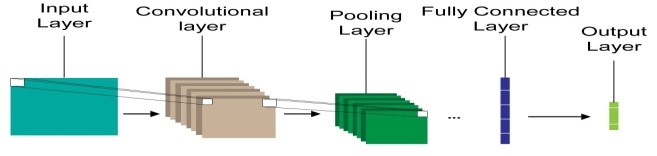}}
\caption{Convolution Neural Network (CNN).}
\label{fig}
\end{figure}
Fully Connected layer exists after the pooling layer. In fully connected. layer nodes have total links to all previous and next layer nodes. From a multi-layer perception, a fully connected layer  plays the character of a classifier \cite{b30}. After performing this layer, we obtain an output.\\
\\
Inception V3 \cite{b31} is a DCNN (Deep Convolution Neural Network) winner of ILSVRC14. Inception V3 is the combination of lots of these blocks shown in Fig. 2. . In Fig. 2, the green-colored rectangles represent the convolution filters or kernels, and the red one is the pooling layers. There are also some extra pooling layers in Inception V3 blocks to change height and weight and softmax regression for making predictions. We used this DCNN for extracting the image functions. We removed the last output layer that the extracted features can be used by GRU for predicting the sentences.
\begin{figure}[!ht]
\centerline{\includegraphics{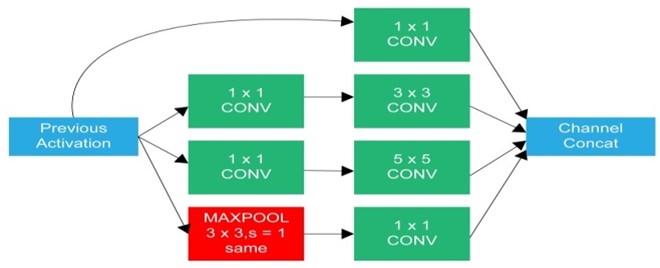}}
\caption{Inception V3 block.}
\label{fig}
\end{figure}\\

\begin{figure}[b]
\centerline{\includegraphics{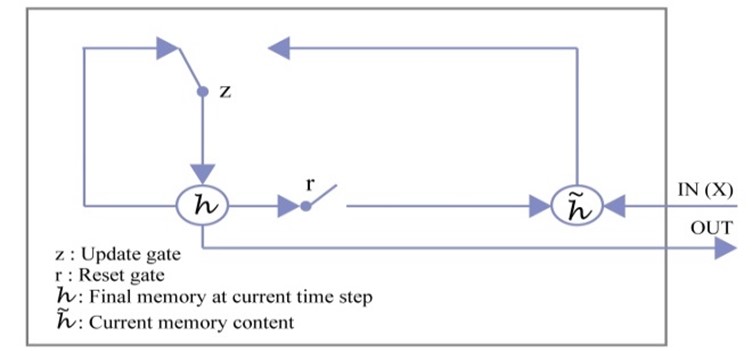}}
\caption{Gated Recurrent Unit (GRU).}
\label{fig}
\end{figure}
Recurrent neural networks (RNN) are a neural network type that is ideally acclimated to process data from sequential and other time-series data. In RNN, the previous step output is fed as the current stage entry. RNN has drastic use of language translation, voice command recognition, speech recognition, sentiment classification, and image caption generation. For these operations, the model learns from inputs with a sequence and provides the outputs with a sequence. \cite{b9}. for the current output, RNN must remember the previous pieces of information for continuous prediction. RNN has memory for remembering all the steps which have been calculated \cite{b32}.\\
\\
Gated Recurrent Unit (GRU) is a particular type of RNN introduced by Cho et al. \cite{b7} in 2014 to subdue the problem of vanishing gradient. The functionalities of GRU is similar to the other famous RNN types LSTM. Unlike LSTM, it has only two gates wherever LSTM has three gates. Because of less gate, GRU has fewer parameters than LSTM. GRU works the same as LSTM with some sequence like (i.e. sound, text). GRU uses so-called reset and update gate to dispel the problem of vanishing gradient. These gates can be trained to keep information for long without losing the information through a period or eliminating unnecessary information to predict \cite{b25}. The whole GRU structure information is shown in Fig. 3.\\
\\
The update gate objective is to assess the model of how much previous information should pass to the future. The formula (1) borrowed from \cite{b25}\\
\begin{equation}
z_t = \sigma (W_zx_t + U_zh_{t-1})                                  
\end{equation}
Here the output of update gate \(z_t\) for t time step. The weight matrix is \(W_z\) of update gate multiplied by the input of the current unit is xt, \(h_{t-1}\) hold the t-1 or previous unit information with the multiplication of \(U_z\), which is its own weight.\\\\
The model uses the reset gate to decide the amount of precise information from the previous units should forget. The formula (2) also borrowed from \cite{b25}
\begin{equation}
r_t = \sigma (W_rx_t + U_rh_{t-1})                                  
\end{equation}
Here the output of reset gate rt for t time step. The weight matrix is Wr of reset gate multiplied by the input of the current unit is \(x_t\), \(h_{t-1}\)hold the t-1 or previous unit information with the multiplication of \(U_r\), which is its own weight.\\\\

The \textbf{current memory content} uses the reset gate to store the information which is relevant from the past. The formula (3) also borrowed from\cite{b25}
\begin{equation}
\widetilde{h_t} =\tanh(W_{xt}+U(r_t \odot h_{t-1}))                       
\end{equation}
Here the output of \textbf{current memory content} \(\widetilde{h_t}\) for t time step. The weight matrix is W multiplied by the input of the current unit is \(x_t\), \(h_{t-1}\) hold the t-1 or previous unit information with the element wise multiplication of \(r_t\) ,\( r_t\) is the output of reset gate and U which is its own weight.\\
The Final memory at current time step sets what to accumulate from the current step and transfer it to the next layer. Update gate is needed for this operation. The formula (4) also borrowed from\cite{b25}
\begin{equation}
h_t=z_t \odot h_t+(1-z_t) \odot \widetilde{h_t}                      
\end{equation}
Here element wise multiplication with update gate zt and information from previous unit \(h_{t-1}\). Where element wise multiplication between \((1- z_t)\) and current unit gives information \(h_t\).\\
\section{New Dataset}
During implementation, for training our model and for generating Bangla captions, we needed a Bangla dataset. We got two Bangla datasets called BANGLALEKHA\cite{b33} and BNLIT\cite{b34} for continued, but those datasets were packed with inaccuracy. They didn't match the standard even though, according to the picture, BNLIT has one caption, and BANGLALEKHA has two captions for each. Because of lack of the dataset, we are proposing our new dataset called BNATURE. Our dataset consists of 8000 pictures with the dimension of 500 x 375 pixels . All these pictures represent Bangladeshi lifestyle and nature. Every image consists of 5 Bengali captions. The structure of the dataset is followed by recognized datasets like flicker8k, flicker30k, MSCOCO\cite{b6}\cite{b19}\cite{b20}\cite{b11}. We used 6000 images for training, 1000 images for testing, and the rest of the images for validation. We used one NVIDIA GTX 1060.\\
\section{Image Preprocessing}
For feeding the dense layer of GRU we produced 2048-D shape. Preprocessing has been done to switch the images into RGB three dimensional arrays. The size of the array is (299, 299, 3). We used pillow, Caffe libraries for optimizations of our pictures.
\section{Caption Preprocessing}
\begin{itemize}
	\item We marked the captions with adding "\<start\>" at first of the captions and at the end of the captions we added "\<end\>" token. This helped the system that it started producing the sentence when it got the "\<start\>" token and it stopped creating the sentence when it saw the "\<end\>" token.
	\item We tokenized by separating the sentence word by word. We also captured all the unique words from our dataset
	\item We converted all the text in sequence to convert all the words into the sequential word. This method produces a vector representing the series of Sentences in terms.
\end{itemize}
\section{Caption Generation}
The sentence generation process in the neural network is taken from the encoder-decoder modeling principle in the network and machine translation modeling\cite{b19}.  Using encoder, the sequence of the word represents a distributed vector in this modeling. Then the model generates a new sequence of words with the experience of CNN by the help of decoder. The main target was to maximize the accuracy of the prediction of the perfect sentence while training. We used Bidirectional Recurrent Neural Network (BRNN) first introduced by Schuster and Paliwal in 1997\cite{b35}, which we used to maximize the accuracy. BRNN predicts each word of the sequence based on previous and future elements. BRNN preform this sequencing using two RNN processing output one perform from left to right and another one from right to left. From these two outputs ultimately we choose the final output of the sequence. The probability of predicting the next word using BRNN inspired from\cite{b36}.
\begin{equation}
\overrightarrow{p}(w_t\mid I)=\prod_{t=1}^Tp(w_t\mid w_1,w_2,w_3…………,w_{t-1},I)
\end{equation}
\begin{equation}
\overleftarrow{p}(w_t\mid I)=\prod_{t=1}^Tp(w_T\mid w_{t+1},w_{t+2},w_{t+3}……… w_T,I)\\
\end{equation}
\begin{equation}
p(w_{1:T}\mid I)=\max(\frac{1}{T} \sum_{t=1}^T(\overrightarrow p(w_t\mid I)),\frac{1}{T}\sum_{t=1}^T(\overleftarrow p (w_t\mid I)))
\end{equation}
where \(\overrightarrow{p}(w_t\mid I)\) predict the word in the way of forward order  on the other hand \(\overleftarrow{p}(w_t\mid I)\) predict the word in the way of backward order with the experience from the previous words. At last we have to choose these final output of sentence from these two order we select more accurate one denoted by \(p(w_{1:T}\mid I)\) the accurate one is selected by the max function. Whereas w is denoted as the one hot vector of 300 dimensional and t is the time step of producing the sentence. I is denoted as the input image. Whereas the loss function is
\begin{equation}
L (I; S)=-\sum_{t=0}^N \log p_t(S_t)
\end{equation}
Beam Search evaluates the collection iteratively of the better k sentences until t as candidates for generating sentences of size t + 1, with only the best outcome k. We used Beam search with beam 3, 5, 7 to generate the captions. We also used argmax, which is considered to give better approximate\cite{b6}.
\section{Evaluation Metrics}
We used Bilingual Evaluation Understudy (BLEU) \cite{b39} and METEOR \cite{b37}for the evaluation of our generated sentence from the model. If the generated sentence from the model matches precisely with the sentence generated by humans, then we will get bleu score of 1.0. If the generated sentence from the model does not match with the sentence made by humans, then we will get 0.0 bleu score. Bleu calculates the accuracy with an n-gram matric and n= {1, 2, 3, and 4}
\begin{equation}
p_n=\frac{\sum_{n-gram\in c }Count_{clip}(n-gram)}{\sum_{n-gram\in c }Count(n-gram)}
\end{equation}
Where c is the model generated output sentence and \(Count \) is the number of n-grams that appear in the output sentence and \(Count_{clip}\)  is the number of n-grams appear in the references according to the output n-gram.
\begin{figure}[!b]
\centerline{\includegraphics[width=1\linewidth,height=1\textheight,keepaspectratio]{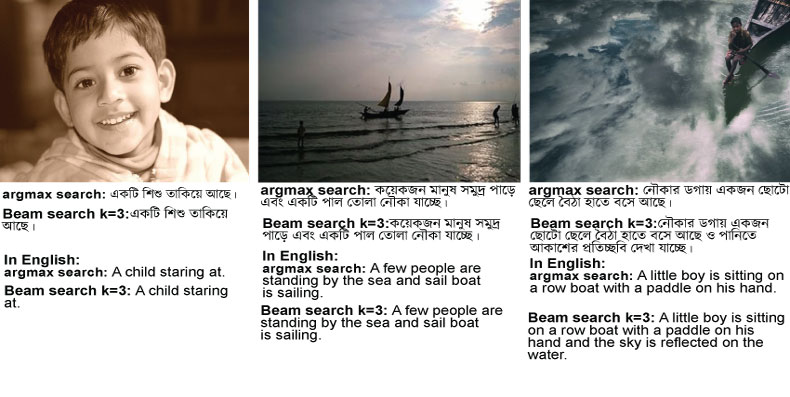}}
\caption{Some samples of system generated sentences according to the pictures.}
\label{fig}
\end{figure}
\section{Result}

We analyze the capacity of the functioning hybrid deep model by looking at how effectively the test photos can be practically represented. We provided 1000 images that are unknown to the model to predict the captions. Every caption generated by the model compared by the 5 other reference sentences which are provided by human evaluation and the comparison score are described by us using Bleu and Meteor score. By using these scores, we could determine if our model is working better or not. \\
In TABLE I, we illustrated the performance of our model using the BLEU and METEOR scores. How well the BEAM search where beam=3 and improved argmax search were shown.

\begin{table}[htbp]
\caption{}
\begin{center}
\begin{tabular}{|c|c|c|c|c|c|}
\hline
\cline{2-4} 
\textbf{\textit{Search}} & \textbf{\textit{BLEU-1}}& \textbf{\textit{BLEU-2}}& \textbf{\textit{BLEU-3}}& \textbf{\textit{BLEU-4}}& \textbf{\textit{METEOR}} \\
\hline
\textbf{BEAM }&42.58&27.95&23.66&16.41&28.70\\
\hline
\textbf{Argmax}&40.54&25.22&20.59&13.50&28.10 \\
\hline
\end{tabular}
\end{center}
\end{table}
After evaluation, we can see the Beam Search with beam 3 perform better than the Argmax Search. Whereas the bleu score and meteor score are 2\% improved in the beam search. 
\begin{figure}[htbp]
\centerline{\includegraphics{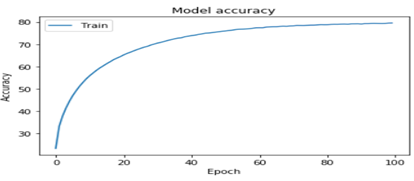}}
\caption{Model performance on Accuracy Vs Epoch.}
\label{fig}
\end{figure}
\begin{figure}[htbp]
\centerline{\includegraphics{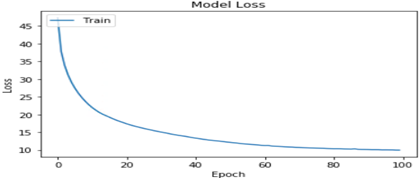}}
\caption{Model performance on Loss Vs Epoch.}
\label{fig}
\end{figure}\\
After training our model up to 100 epochs, our accuracy has been improved from 10\% to 8\%, then It got plateau shown in Fig. 5, and our loss has been decreased from 50\% to\% shown in Fig. 6.
\section{Discussion}
After testing the trained model with 1000 pictures, the bleu score show in Table1 is satisfactory. Our new dataset helps us to achieve this result. In Fig. 4, we can see some pictures and model generated sentences based on the type of searching algorithm. We translated the model generated caption for those who do not know the Bengali language for them. In the first picture, we can see the model can detect a kid or an adult's photo. In the second one, the generated sentence is perfect according to the picture, and in the last one, it also can detect in which part of the boat the kid is sitting at, and even there is a reflection of the sky in the water. The generated sentence is quite detailed and accurate. Most of the model-generated captions are perfect or have little errors, but few of them are thoroughly irrelevant according to the test images. We evaluate our model with BELU=1, 2, 3, 4-gram metrics. Beam search performed better than argmax in every n-gram. On the other hand, the meteor score between the beam and argmax search has not many differences. Although our mRNN (multimodal Recurrent Neural Network) performed flawlessly, the limitation is that the Recurrent Neural Network can obtain the data of all pictures only by additive bias intercommunications, which are inadequate articulate than more intricate multiplicative interactions.
\section{Conclusion}
In conclusion, there are three significant benefits in the three sectors. Firstly, blind peoples are taking benefits from it. Secondly, Automated Robot can also benefit from it. Finally, it will be an excellent system for foreigners who want to visit Bangladesh and want to know Bangladesh's culture. That means this system will play a significant role in Bangladesh's tourism sector because this system can generate the Bengali caption of any pictures. Deep learning is a large area of machine learning. We use some bit of it. In the future, many systems can be made for humanity. By doing this, we keep a sign for making the Bangladesh Digital. We have used CNN (Convolution Neural Network) and RNN (Recurrent Neural Network) for producing our captions. For reliability and improved outcomes, we replaced the LSTM layer with GRU. Furthermore, it removes the vanishing gradients problem. We have created a new dataset with 8000 pictures because of the error in existing datasets, and every picture has five captions. We used the pillow, Caffe libraries for optimizations of our photographs. We used Beam search with beam 3, 5, 7 to generate the captions. We also used argmax, which is considered to give a better approximate. For evaluating our system, we measure the bleu score and meteor score. Beam Search with beam 3 helped to improve our results. This system can play a significant role in the automation sector in Bangladesh. We got an accuracy of 80\%, but if the dataset becomes more massive, there are more chances to increase the accuracy. In future work, the bigger dataset of Bangla can improve the accuracy and the BLEU score.

\end{document}